\documentstyle[aclap]{article}

\title{Adapting the Core Language Engine to French and Spanish}

\author{Manny Rayner and David Carter\\ 
SRI International \\ 
Suite 23, Millers Yard\\ 
Cambridge CB2 1RQ\\ 
United Kingdom \\ 
{\tt \{manny, dmc\}@cam.sri.com}
\And 
Pierrette Bouillon \\ 
ISSCO, University of Geneva\\ 
54, route des Acacias\\ 
1227 Geneva\\ 
Switzerland\\ 
{\tt Pierrette.Bouillon@issco.unige.ch}} 

\begin{document}

\maketitle
\vspace{-0.5in}

\begin{abstract}

We describe how substantial domain-independent language-processing
systems for French and Spanish were quickly developed by manually
adapting an existing English-language system, the SRI Core Language
Engine. We explain the adaptation process in detail, and argue that it
provides a fairly general recipe for converting a grammar-based system
for English into a corresponding one for a Romance language.

\end{abstract}

\bibliographystyle{acl}

\section{Introduction}

In this paper, we will describe how substantial domain-independent
language-processing systems for French and Spanish were quickly
developed by manually adapting an existing English system, the SRI
Core Language Engine. The resulting systems have been integrated as
components in the Spoken Language Translator (SLT;
\cite{SLT-HLT,SLT-report}). The English to French version of SLT
\cite{RaynerBouillon:95} is of a standard comparable to the original
English to Swedish version.

The syntactic rule-set for French covers nearly all the basic
constructions of the language, including the following: declarative,
interrogative and imperative clauses; formation of YN and WH-questions
using inversion, complex inversion and ``est-ce que''; clitic
pronouns; adverbial modification; negation; nominal and verbal PPs;
complements to ``\^{e}tre'' and ``il y a''; relative clauses,
including those with ``dont''; partitives, including use of 
``en''; passives; pre- and post-nominal adjectival modification,
including comparative and superlative; code expressions; sentential
complements and embedded questions; complex determiners; numerical
expressions; date and time expressions; conjunction of most major
constituents; and a wide variety of verb types, including modals and
reflexives. There is a good treatment of inflectional morphology which
includes all major paradigms.  The coverage of the Spanish grammar is
comparable in scope, though slightly less extensive.  The French and
Spanish versions of the CLE are both ``reversible'', and can be used
for either analysis or generation.

We will describe the adaptation process in detail, and argue that it
provides a fairly general recipe for converting a grammar-based system
for English into a corresponding one for a Romance language.  Due to
space limitations, and since it is rather the better of the two, we
will concentrate on the French version. Examples will be taken
from the Air Travel Planning (ATIS) domain used in the current
SLT prototype.

The rest of the paper is organized as
follows. Section~\ref{Section:CLE} gives an overview of the CLE,
focussing on the aspects relevant to this paper.
Section~\ref{Section:Morphology} describes the French morphology
rules. Sections~\ref{Section:French} and~\ref{Section:Spanish}
describe the French and Spanish
grammars. Section~\ref{Section:Conclusions} concludes.

\section{Overview of the Core Language Engine}
\label{Section:CLE}

The CLE is a general language-processing system, which has been
developed by SRI International in a series of projects starting in
1986. The original system was for English only.  A Swedish
version~\cite[\S14.2]{CLE} 
was developed in a collaboration with
the Swedish Institute of Computer Science; the French and Spanish
versions described here were developed in collaborations with ISSCO
and the University of Seville respectively. The CLE is extensively
described elsewhere \cite{CLE,CLARE-report,SLT-report}, so this
section will only give the minimum background necessary to understand
the remainder of the paper.

The basic functionality offered by the CLE is two-way translation
between surface form and a representation in terms of a logic-based
formalism called QLF \cite{AlshawiCrouch:92}. The modules comprising
a version of the CLE for a given language can be divided into three
groups, which we refer to as ``code'', ``rules'' and
``preferences''. The ``code'' modules constitute the
language-independent compilers and interpreters that make up the basic
processing engine; the other two types of module between them
constitute a declarative description of the language.

The ``rules'' contain domain-independent lexico-grammatical
information for the language in question; they encode a
relationship between surface strings and QLF representations. Thus for
any given surface string, the rules define a set of possible QLF
representations of that string.  Conversely, given a well-formed QLF
representation, the rules can be used to produce a set of possible
surface-form realisations of the QLF. The code modules support
compilation of the rules into forms that allow fast processing in both
directions: surface-form $\rightarrow$ QLF (analysis) and QLF
$\rightarrow$ surface-form (generation).

The relationship between surface form and QLF is in general
many-to-many. ``Preference'' modules contain data in the form of
statistically learned distributional facts, based on analysis of
domain corpora \cite{AlshawiCarter:94,RaynerBouillon:95}).  Using
this extra information, the system can distinguish between plausible
and implausible applications of the rules with a fairly high degree of
accuracy. In particular, the preference information makes it possible
in a given application domain to select the intended readings of
ambiguous utterances. We will not consider the preference modules
further here, as the statistical training procedures are completely
language-independent and invisible to the developer.

The non-trivial problems involved in adapting the CLE to a 
new language arise in connection with the ``rule'' modules,
which we will now consider in more detail. These fall into
the following main categories:
\begin{description}
\item[Lexicon:] The largest single set of rules is that which
comprises the lexicon. This is in fact divided up into three subsets:
a function-word lexicon; a set of macros specifying generic lexical
entries for common types of content word (e.g.\ ``count noun'',
``intransitive verb'', etc.); and a content word lexicon which defines
lexical entries for other words in terms of the generic macros.
\item[Morphology:] This set of rules defines the inflectional morphology
of the language, allowing analysis of words in terms of stems
and inflections. The morphology rules are described further in
Section~\ref{Section:Morphology}.
\item[Syntax:] Syntax rules are written in a uni\-fi\-cation-based
feature-grammar formalism. The style of the
CLE grammar is loosely based on GPSG; detailed descriptions
of the CLE grammar for English are available
in~\cite{CLE-grammar-chapter} and~\cite{SLT-grammar-chapter}.
\item[Semantics:] The CLE grammar is
``sign-based''~\cite{CLE-semantics-chapter}; each syntax rule is
coupled with one or more semantic counterpart, which defines the piece
of QLF form produced by that rule. QLF representations are built up
compositionally using unification only.
\item[Reference resolution and scoping:] Further sets of
rules can be used to convert QLF representations into 
representations in full first-order logic. This phase of
processing is required, for example, when using the CLE
for database query applications~\cite{AlshawiCrouch:92,Rayner:93}
\end{description}
As noted in~\cite[\S14.2.2]{CLE}
the effort involved in adapting a set of rule modules to a new
language depends on how directly they refer to surface form;
unsurprisingly, modules defining surface phenomena are the ones which
require most work. When adapting the system to French and Spanish, the
problems arose almost exclusively in connection with morphology and
syntax rules. Other parts of the English system were adapted with
little effort. In particular, the semantic rule-sets for English could
be used for the new languages with only minimal changes.

The following two sections describe in detail the issues pertaining 
to the morphology and syntax rule-sets respectively.

\section{Morphology and spelling}
\label{Section:Morphology}

In order to handle the more complex inflectional morphology of Romance
and other European languages, a morphological processor based on
feature-augmented two-level morphology was developed
\cite{Carter:95}. This allows the complex spelling changes occurring
in these languages to be handled quickly in both analysis and
generation.  Compilation of the full sets of two-level rules
describing spelling changes and of production rules describing legal
affix combinations takes of the order of a minute, allowing changes to
the rules to be debugged relatively easily. Further flexibility is
gained by not requiring the lexicon to be present at compile time
(contrast \cite{KaplanKay:94}); thus the lexicon can be incremented
and tested without any recompilation being required. Two-level
spelling rules were also used to describe the inter-word effects that
are particularly common in French.

The total number of rules required to describe inflectional morphology
was around 75 for French and 50 for Spanish (inter-word rules being
responsible for much of the difference). 
We concentrate here on the French phenomena, which are more complex.

\subsection{Intra-word spelling changes}

Intra-word spelling changes for French present several problems not
encountered in English inflectional morphology. Some of these are
technical in nature, and easily dealt with. In particular, French
exhibits many multiple letter changes, e.g.\ ``chameau+e''
$\rightarrow$ {\it chamelle}, ``peign+rai'' $\rightarrow$ {\it
peindrai}. For reasons explained in \cite{Carter:95}, these must be
handled by a separate rule for each letter that changes, rather than
one for the whole changed substring. Also, some changes can be
optional. For example, the ``y'' in verbs such as ``payer'' can remain
the same or change to ``i'' before silent ``e'': ``pay+e'' $\rightarrow$
either {\it paye} or {\it paie}. This phenomenon is rare or absent in
English, but is handled easily by making the relevant spelling rule
optional.

Less trivial problems, however, arise from the fact that spelling
changes in French generally cannot be predicted from the surface form
of the word alone. This means the application of the rules must be
controlled; we do this by specifying feature constraints, which must
match between the rule and all morphemes it applies to. The following
extended example describes our treatment of one of the most
challenging cases.

Nouns, adjectives and verbs ending in ``-et'' or ``-el''
can either double the ``t'' or ``l'' before a silent ``e'' or change
the prefinal ``e'' to ``\`{e}'': ``cadet+e'' $\rightarrow$ {\it
cadette}, but ``complet+e'' $\rightarrow$ {\it compl\`{e}te}.  The
application of the spelling rules is therefore controlled by means of
a feature {\tt spelling\_type}, with value {\tt double} in the first
case and {\tt change\_e\_\`{e}} in the second.

This situation is further complicated by two facts. Firstly, the surface 
``\`{e}l'' or ``\`{e}t'' of the verbs is ambiguous between a deep 
``el'' or ``et'', and ``\'{e}l'' or ``\'{e}t''. For example, 
we have {\it ach\`{e}te} $\leftarrow$ ``achet+e'', but 
{\it affr\`{e}te} $\leftarrow$ ``affr\'{e}t+e''. For this reason, we 
introduce a third value for {\tt spelling\_type}: 
{\tt change\_\'{e}\_\`{e}}. ``Affr\'{e}t'' has thus the feature 
{\tt spelling\_type=change\_\'{e}\_\`{e}}, ``achet'' 
{\tt spelling\_type=change\_e\_\`{e}} and ``appel'' 
{\tt spelling\_type=double}.

Secondly, the ``e'' that begins future and conditional endings sometimes 
affects preceding letters as if it were silent, and sometimes as if it 
were not.  For example, ``appel+erai'' $\rightarrow$ {\it appellerai}, 
doubling the ``l'' just as in ``appel+e'' $\rightarrow$ {\it appelle}, 
where the final ``e'' actually is silent. However, ``c\'{e}d+erai''
$\rightarrow$ {\it c\'{e}derai}, not *{\it c\`{e}derai} as would be
expected from the silent-e behaviour ``c\'{e}d+e'' $\rightarrow$ {\it
c\`{e}de}.  To make this distinction, we use a feature {\tt muet}
(``silent'') for specifying if the ``e'' in the suffix is silent, 
as ``e'' ({\tt muet=y}), not silent, as ``ez'' ({\tt muet=n}) 
or the ``e'' of the future or conditional tenses, for example ``erai/erais'' 
({\tt muet=fut\_cond\_e}). Then, we restrict the rule for doubling
the consonant with the features {\tt spelling\_type=double, 
muet=y$\vee$fut\_cond\_e}, and the one for ``\'{e}'' 
$\rightarrow$ ``\`{e}'' with the features 
{\tt spelling\_type=change\_\'{e}\_\`{e},muet=y}. 

\subsection{Inter-word spelling changes}

In English, inter-word spelling changes occur only in the alternation
between ``a'' and ``an'' before consonant and vowel sounds
respectively. In French, such changes are far more widespread and can
be complex.  However, they can be handled by judiciously specifying
contexts in two-level rules and, in a few cases, by postulating
non-obvious underlying lexical items.  Some important cases are:
\begin{itemize}
\item The ``e'' in the function words ``de'', ``je'', ``le'', ``me'',
``ne'', ``que'', ``se'' and ``te'' is elided before (most) words
starting in a vowel sound, except when the function word follows a
hyphen: ``le homme'' $\rightarrow$ {\it l'homme}, ``je ai''
$\rightarrow$ {\it j'ai}, but ``puis-je avoir'' does not elide, so the
elision rule specifies that the hyphen be absent from the
context. ``Ce'' also elides when used as a pronoun (``ce est''
$\rightarrow$ {\it c'est}, but when used as a determiner it takes the
form ``cet'' before a vowel: {\it cet homme}. We therefore take the
underlying form of the determiner to be ``cet'', which {\it loses} its
``t'' when followed by a consonant-initial word (``cet soir''
$\rightarrow$ {\it ce soir}).  

Numerals do not allow elision either: ``le onze'' does not become
*{\it l'onze}. We therefore treat the lexical form as being
``\#onze'', where ``\#'' acts as an underlying consonant but is
realised as a null.  (Syntax plays a role here too: ``le un''
$\rightarrow$ {\it l'un} when is a determiner, but not when it is a
numeral. Thus lexically we have ``un'' as determiner and ``\#un'' as
numeral).

\item The very common preposition/article combinations
``de''/``\`{a}'' and ``le''/``les'': ``de le'' $\rightarrow$ {\it du},
``\`{a} les'' $\rightarrow$ {\it aux}, etc.  These contractions span
constituent boundaries (we view {\it du vol} as being syntactically
[PP de [NP le vol]]) so need to be treated as spelling effects. Also,
vowel elision takes precedence: ``de le homme'' $\rightarrow$ {\it de
l'homme}, not *{\it du homme}.

\item Hyphens between verbs and clitic pronouns are treated as lexical
items in our grammar. They are realised as {\it -t-} when preceded by
``a'' or ``e'' and followed by ``e'', ``i'' or ``o'': ``va - il''
$\rightarrow$ {\it va-t-il}, but ``vont - ils'' $\rightarrow$
{\it vont-ils}. Hyphens joining nouns or names are treated as different
lexical items not subject to this change: ``les vols Atlanta -
Indianapolis'' does not involve introduction of ``t''.
\end{itemize}

\section{French syntax}
\label{Section:French}

When comparing the French and English grammars, there are two types of
objects of immediate interest: {\it syntax rules} and {\it features}.
Looking first at the rules themselves, about 80\% of the French syntax
rules are either identical with or very similar to the English counterparts
from which they have been adapted. Of the remainder, some rules (e.g\
those for date, time and number expressions) are different, but
essentially too trivial to be worth describing in detail. Similar
considerations apply to features. 

We will concentrate our exposition on the rules and features which are
both significantly different, and possess non-trivial internal
structure. Examining the grammar, we find that there are three large
interesting groups of rules and features, describing three separate
complexes of linguistic phenomena: question-formation, clitic pronouns
and agreement. As we have argued previously~\cite{RaynerBouillon:95},
all of these are rigid and well-defined types of construction which
occur in all genres of written and spoken French. It is thus both
desirable and reasonable to attempt to encode them in terms of
feature-based rules, rather than (for instance) expecting to derive
them as statistical regularities in large corpora.
In Sections~\ref{Section:Clitics},~\ref{Section:Questions}
and~\ref{Section:Agreement}, we describe how we handle these key
problems.

\subsection{Question-formation}
\label{Section:Questions}

We start this section by briefly reviewing the way in which
question-formation is handled in the English CLE grammar. There are
two main dimensions of classification: questions can be either WH- or
Y-N; and they can use either the inverted or the uninverted
word-order. Y-N questions must use the inverted word-order, but both
word-orders are permissible for WH-questions. The phrase-structure
rules analyse an inverted WH-question as constituting a fronted WH+
element followed by an inverted clause containing a gap element.  The
feature {\tt inv} distinguishes inverted from uninverted clauses.
The following examples illustrate the top-level structure of
Y-N, unmoved WH- and moved WH-questions respectively.
\begin{quote}
   [Does he love Mary]$_{S:[inv=y]}$ 
\end{quote}
\begin{quote}
   [Who loves Mary]$_{S:[inv=n]}$
\end{quote}
\begin{quote}
   [[Whom]$_{NP}$ [does he love []$_{NP}$]$_{S:[inv=y]}$]
\end{quote}

The French rules for question formation are structurally fairly similar
to the English ones. However, there are several crucial differences which
mean that the constructions in the two languages often differ widely
at the level of surface form. Two phenomena in particular stand out.
Firstly, English only permits subject-verb inversion when the verb
is an auxiliary, or a form of ``have'' or ``be''; in contrast,
French potentially allows subject-verb inversion with any verb.
For this reason, English question-formation using auxiliary ``do''
lacks a corresponding construction in French.

Secondly, French permits two other common question-formation
constructions in addition to subject-verb inversion: prefacing the
declarative version of the clause with the question particle ``est-ce
que'', and ``complex inversion'', i.e.\ fronting the subject and inserting a
dummy pronoun after the inverted verb. In certain circumstances,
primarily if the subject is the pronoun ``\c{c}a'', it is also
possible to form a non-subject WH-question out of a fronted WH+ phrase
followed by an uninverted clause containing an appropriate gap. We
refer to this last possibility as ``pseudo-inversion''.

If the subject is a pronoun, only inversion and the ``est-ce que''
construction are allowed; if it is {\it not} a pronoun, only the
``est-ce que'' construction and complex inversion are valid. In
addition, a subject pronoun following an inverted verb needs to be
linked to it by a hyphen, which can be realised as a ``-t-''
(cf. Section~\ref{Section:Morphology}).  
Figure~\ref{Figure:FrenchQ}
presents examples illustrating the main French question
constructions.

\begin{figure}
\begin{quote}
{\it Y-N, inversion:}\\
Aime-t-il Marie?\\
\\
{\it Y-N, ``est-ce que'':}\\
Est-ce que Jean aime Marie?\\
\\
{\it Y-N, complex inversion:}\\
Jean aime-t-il Marie?\\
\\
{\it WH, subject question, no inversion:}\\
Quel homme aime Marie?\\
\\
{\it WH, inversion:}\\
Quelle femme aime-t-il?\\
\\
{\it WH, ``est-ce que'':}\\
Quelle femme est-ce que Jean aime?\\
\\
{\it WH, complex inversion:}\\
Quelle femme Jean aime-t-il?\\
\\
{\it WH, pseudo-inversion:}\\
Combien \c{c}a co\^{u}te?
\end{quote}
\caption{Main French question constructions}
\label{Figure:FrenchQ}
\end{figure}

Modification of the English syntax rules to capture the basic
requirements so far is quite simple. In our grammar, we added three
extra rules to cover the ``est-ce que'', complex-inversion and
pseudo-inversion constructions: the second of these rules combines the
complex-inverted verb with the following dummy pronoun to form a verb,
in essence treating the dummy pronoun as a kind of verbal affix. A
further rule deals with the hyphen linking an inverted verb with a
following subject.

With regard to the feature-set, the critical change involves the {\tt
inv} feature. In English, as we saw, this feature had two possible
values, {\tt y} and {\tt n}. In French, the corresponding feature has
five values: {\tt inverted}, {\tt uninverted}, {\tt est\_ce\_que},
{\tt complex} and {\tt pseudo}, distinguishing clauses formed using
the different question-formation constructions. (It is important to
note, though, that the semantic representation of the clause is the
same irrespective of its inversion-type). To enforce the restrictions
concerning combinations of inversion-type and subject form, we also
added a new clausal feature which distinguished clauses in which the
subject is a pronoun. 

The attractive aspect of this treatment is that the remaining English
rules used for question-formation can be retained more or less
unchanged. In particular, the English semantic rules can still be
used, and produce QLF representations with similar form. 

It would almost be true to claim that the above constituted our 
entire treatment of French question-formation. In practice, we
have found it desirable to add a few more features to the grammar
in order to block infelicitous combinations of the inversion
rules with certain commonly occurring lexical items. It is 
possible that the effect of these features could be achieved 
equally well by statistical modelling or other means, but
we describe them here for completeness:
\begin{description}
\item[Restrictions on use of ``est-ce que'':] Question-formation with
``est-ce que'' is strongly dispreferred when the main verb is a
clause-final occurrence of ``\^etre'', or existential ``avoir'' (as in
``il y a''). For example:
\begin{quote}
?Quand est-ce que le prochain vol est?\\
?Combien de vols est-ce qu'il y a?
\end{quote}
We enforce this by adding a suitable feature to the verb
category. 
\item[Fronting of ``heavy'' NPs:] Most languages prefer not to
front ``heavy'' NPs, and this dispreference is particularly
strong in French. We have consequently added an NP feature 
called {\tt heavy}, which has the value {\tt y}
on NPs containing PP and VP post-modifiers. Thus for example
generation of
\begin{quote}
Quels vols en partance de Dallas y a-t-il?
\end{quote}
is blocked, but the preferable
\begin{quote}
Quels vols y a-t-il en partance de Dallas?
\end{quote}
is permitted.
\item[Inverted subject NPs:] Occurrence of some pronouns (in
particular ``cela'', and ``\c{c}a'') is strongly dispreferred in
inverted subject position.
A binary feature enforces this as a rule, for example
blocking
\begin{quote}
Combien co\^{u}te \c{c}a pour aller \`{a} Boston?
\end{quote}
but instead permitting
\begin{quote}
Combien \c{c}a co\^{u}te pour aller \`{a} Boston?
\end{quote}
\end{description}

\subsection{Clitics}
\label{Section:Clitics}

The most difficult technical problems in adapting an English grammar
to a Romance language are undoubtedly caused by clitic pronouns. In
contrast to English, certain proform complements of verbs do not
appear in their normal positions; instead, they occur adjacent to the
main verb, and possibly joined to it by a hyphen. The position of the
clitics in relation to the verb (pre- or post-verbal) is determined by
the mood of the verb, and whether or not the verb is negated. If
two or more clitics are affixed to the verb, their internal order is
determined by their surface forms. Several attempts to account for the
above and other data have previously been described in the literature
e.g.\ \cite{Grimshaw:82,BesGardent:89,Estival:90,MillerSag:95}; we
have in particular been influenced by the last of these,.

Although the underlying framework is very different from the HPSG
formalism used by Miller and Sag, our basic idea is the same: to treat
``clitic movement'' by a mechanism similar to the one used to handle
WH movement. More specifically, we introduce two sets of new
rules. The first set handles the ``surface'' clitics. They define the
structure of the verb/clitic complex, which we, like Estival, regard
as a constituent of category V composed of a main verb and a
``clitic-list''. A second set of ``gap'' rules defines empty
constituents of category NP or PP, occurring at the notional ``deep''
positions occupied by the clitics.  Thus, for example, on our account
the constituent structure of ``Est-ce que vous le voulez?'' will be
\begin{quote}
[Est-ce que [vous$_{NP}$ [le voulez]$_V$ []$_{NP}$]$_S$]$_S$
\end{quote}
where the ``gap'' NP category represents the notional direct
object of ``voulez'', realised at surface level by the pre-verbal
clitic ``le''.

To make this work, we add an extra feature, {\tt clitics}, to all
categories which can participate in clitic movement: in our grammar,
these are V, VP, S, NP and PP. The {\tt clitics} feature is used
to link the cliticised V constituent and its associated clitic gap
or gaps. We have found it convenient to define the value of the 
{\tt clitics} feature to be a bundle of five separate sub-features,
one for each of the five possible clitic-positions in French. 
Thus for instance the second-position clitics ``le'', ``la'' and
``les'' are related to object-position clitic gaps through the
second sub-feature of {\tt clitics}; the fourth-position ``y''
clitic is related to its matching PP gap through the fourth
sub-feature; and so on. 
The linking relation between a clitic-gap
and its associated clitic is formally exactly the same as
that obtaining between a WH-gap and its associated antecedent,
and can if desired be conceptualized as a type of coindexing.

The {\tt clitics} feature-bundle is threaded through the grammar rule
which defines the structure of the list of clitics associated with
a cliticised verb, and enforces the constraints on ordering of
surface clitics. These constraints are encoded in the lexical
entry for each clitic. 

This basic framework is fairly straight-forward, though a number of
additional features need to be added in order to capture the
syntactic facts. We summarize the main points:
\begin{description}
\item[Position of surface clitics:] Clitics occur post-verbally in
positive imperative clauses, otherwise pre-verbally. The clitic-list
constituent consequently needs to share suitable features with the
verb it combines with. 
\item[Surface form of clitics:] The first- and second-person singular
clitics are realised differently depending on whether they occur pre-
or post-verbally: for example ``Vous me r\'{e}servez un vol'' versus
``R\'{e}servez-moi un vol''. Moreover, ``me'' and ``te'' are
first-position clitics (e.g.\ ``Vous me les donnez''), while ``moi''
and ``toi'' are third-position (``Donnez-les-moi''). This alternation
is achieved simply by having separate lexical entries for each
form. The entries have different syntactic features, but a common
semantic representation.
\item[Special problems with the ``en'' clitic:] The most abstruse
problems occur in connection with the ``en'' clitic, and are
motivated by sentences like
\begin{quote}
Combien en avez-vous?
\end{quote}
Here, our framework seems to dictate a constituent structure including
three gaps, viz:
\begin{quote}
[Combien [[en avez]$_V$ [vous$_{NP}$ [[]$_V$ [[]$_{NP}$ []$_{PP}$]$_{NP}$]]$_S$]$_S$]$_S$
\end{quote}
in which the V gap links to ``avez'', the NP gap to ``combien'', and
the PP gap to ``en''. The specific difficulty here is that the ``en''
PP gap ends up as an NP modifier (it modifies the NP gap). Normally,
however, PP modifiers of NPs cannot be gaps, and the above type of
construction is the only exception we have found.

Rather than relax the very common {\tt NP $\rightarrow$ NP PP} rule to
permit a gap PP daughter, we introduce a second rule of this type
which specifically combines certain NPs, including suitable gaps
resulting from WH-movement, and an ``en'' clitic gap. A feature, {\tt
takespartative}, picks out the NPs which can participate as left
daughters in this rule.
\end{description}

\subsection{Agreement}
\label{Section:Agreement}

Although grammatical agreement is a linguistic phenomenon that plays a
considerably larger role in French than in English, the adjustments
needed to the lexicon and syntax rules are usually obvious. For
instance, a feature has to be added to the both daughters of the rule
for pre-nominal adjectival modification, to enforce agreement in
number and gender. In nearly all cases, this same procedure is used. A
feature called {\tt agr} is added to the relevant categories, whose
value is a bundle representing the category's person, number and
gender, and the {\tt agr} feature is shared between the categories
which are required to agree.

There are however some instances where agreement is less trivial.  For
example, the subject and nominal predicate complement of ``\^etre'' may
occasionally fail to agree in gender, e.g.
\begin{quote}
La gare est le plus grand b\^atiment de la ville.
\end{quote}
However, if the predicate complement is a pronoun (``lequel'',
``celui-ci'', ``quel''\footnote{Most French grammars regard ``quel''
as an adjective, but for semantic reasons we have found it more
convenient to treat it as a pronoun in this type of construction and
as a determiner in expression like ``quel vol''.}...)  agreement in
both gender and number is obligatory: thus for instance
\begin{quote}
Quel/*quelle/*quels est le premier vol.
\end{quote}
It would be most unpleasant to duplicate the syntax rules, with
separate versions for the pronominal and non-pronominal cases.
Instead, we add a second agreement feature ({\tt compagr}) to the NP
category, which is constrained to have the same value as {\tt agr} on
pronominal NPs; subject/predicate agreement can then use the 
{\tt compagr} feature on the predicate, getting the desired behaviour.

Similar considerations apply to the rule allowing modification of
a NP by a ``de'' PP. In general, there is no requirement on 
agreement between the head NP and the NP daughter of the PP. However,
for certain pronominal NP (``lequel'', ``l'un'', ``chacun'')
gender agreement is obligatory, e.g.
\begin{quote}
lequel/*laquelle de ces vols  \\
laquelle/*lequel de ces dates
\end{quote}
This is dealt with correspondingly, by addition of a new agreement
feature specific to the {\tt NP $\rightarrow$ NP PP} rule.

\section{Spanish syntax}
\label{Section:Spanish}

This section briefly describes the interesting features of the Spanish
syntactic rule-set. In general, the Spanish rules were distinctly
simpler than the French ones.  With a few exceptions noted below (in
particular, prodrop), the current Spanish syntax rules are essentially
a slightly modified subset of the French ones. Despite this, they give
very adequate coverage of the ATIS domain, the only in which they have
so far been seriously tested. In a little more detail:
\begin{description}
\item[Question-formation:] The Spanish rules for question-formation
are similar to, but less elaborate than the French ones. Subject-verb
inversion is allowed with any subject; there is no restriction that it
be pronominal. There are no constructions corresponding to ``est-ce
que'' or complex inversion. When the inverted subject is a pronoun, it
does not require a preceding hyphen linking it to the verb.
\item[Clitics:] The Spanish clitic system is also considerably simpler
than the French one. There are fewer clitics; in particular, there 
are no clitics corresponding to the French ``y'' and ``en'', which as
we saw in Section~\ref{Section:Clitics} above gave rise to many of the
difficult problems in French. 

Postverbal clitics are affixed directly to the verb, rather than being
joined by hyphens. Since CLE morphotax rules have a uniform
format \cite[\S 3.9]{CLE}, this only involved moving the
relevant syntax rules to the morphology rule file. 
\item[Phrasal rules:] The rules for Spanish numbers, dates and times
are substantially different from the French ones, and those for dates
in particular needed to be rewritten more or less from scratch. The
issues involved are however straight-forward.

Also, the form of the Spanish superlative adjective is slightly
different: the postnominal superlative adjective has no extra article,
e.g.\ ``le vol le [plus cher] versus ``la plaza [menos cara]''.
The necessary adjustments are again simple.
\item[Relative clauses:] A less trivial difference involves
relative clauses. In Spanish, the main verb of
the relative clause must be in the subjunctive mood if it 
modifies an argument of a verb in the imperative mood. Thus for example
\begin{quote}
     Which is the first flight that serves a meal?  \\
     $\rightarrow$ Cu\'{a}l es el primer vuelo que sirve una comida?
\end{quote}
(``sirve'' = present indicative), but
\begin{quote}
     Show me flights that serve a meal! \\
     $\rightarrow$ Ens\'{e}\~{n}eme los vuelos que sirva una comida
\end{quote}
(``sirva'' = present subjunctive). Handling this alternation
correctly involves trailing an extra feature through many grammar
rules, so as to link the main verb in the relative clause to
the main verb in the clause immediately above it.
\item[Prodrop:] The second substantial change required when adapting
the French grammar to Spanish was necessitated by the prodrop rule:
Spanish, unlike French, permits and indeed encourages omission of the
subject when it is a pronoun. Perhaps surprisingly, prodrop in fact
only resulted in a few divergences between the Spanish and French
grammars. A new syntax rule of the form {\tt S $\rightarrow$ VP} was
added (it is in fact a slightly modified version of the French
imperative-formation rule). The associated semantic rule fills in a
representation of the omitted clausal subject from the main verb; to
make this possible, the semantic entries for inflected verbs are all
modified to contain an extra feature encoding the possible prodrop
subject. The details are straight-forward.
\end{description}

\section{Conclusions}
\label{Section:Conclusions}
The preceding sections describe in essence all the changes we needed
to make in order to adapt a substantial English language processing
system to French and Spanish. Due to space limitations, we have been
obliged to present some of the details in a more compressed form than
we would ideally have wished, but nothing important has been
omitted. Creation of a good initial French version required about five
person-months of effort; after this, the Spanish version took only
about two person-months. We do not believe that we were greatly aided
by any special features of the Core Language Engine, other than the
fact that it is a well-engineered piece of software based on sound
linguistic ideas. Our overall conclusion is that an English-language
system conforming to these basic design principles should in general
be fairly easy to port to Romance languages.

\section*{Acknowledgements}

The work described here was supported by SRI International, Suissetra,
and Telia Research AB, Sweden. We would like to acknowledge the
assistance provided by Gabriela Fernandez of the University of Seville
in developing the Spanish version of the system, and thank Sabine
Lehmann, David Milward and Steve Pulman for helpful comments.

\end{document}